\documentclass[lettersize,journal]{IEEEtran}
\usepackage{amsmath,amsfonts}
\usepackage{amssymb}
\usepackage{amsthm}
\usepackage[capitalize,noabbrev]{cleveref}
\usepackage{caption}
\usepackage{subcaption}
\usepackage{multirow}
\usepackage{booktabs}
\usepackage{algorithmic}
\usepackage{algorithm}
\usepackage{array}
\usepackage[caption=false,font=normalsize,labelfont=sf,textfont=sf]{subfig}
\usepackage{multirow}
\usepackage{multicol}
\usepackage{verbatim}
\usepackage{longtable}
\usepackage{supertabular}
\usepackage{float}
\usepackage{enumitem}
\usepackage{tablefootnote}
\usepackage{xcolor}
\usepackage{xspace}
\usepackage{textcomp}
\usepackage{makecell}
\usepackage{lscape} 
\usepackage{siunitx}
\usepackage{xfrac}
\usepackage{setspace}
\usepackage{wrapfig}
\usepackage{tabulary}
\usepackage{tabularx}
\usepackage[export]{adjustbox}
\usepackage{threeparttable}
\usepackage{enumitem}
\usepackage{tablefootnote}
\usepackage{textcomp}
\usepackage{stfloats}
\usepackage{url}
\usepackage{verbatim}
\usepackage{graphicx}
\usepackage{cite}
\usepackage{xco lor}

\usepackage{pgfplots}
\usetikzlibrary{pgfplots.groupplots}
\pgfplotsset{compat=1.3}
\usepackage{tikz}
\usetikzlibrary{patterns}
\usepackage{pifont}
\newcommand{\circlednumber}[1]{\scalebox{1.2}{\ding{\numexpr171+#1}}}

\newcommand*\circled[1]{\tikz[baseline=(char.base)]{
            \node[shape=circle,draw,inner sep=0.5pt] (char) {#1};}}

\newcommand{\greendown}{\textcolor{green}{$\downarrow$}}
\newcommand{\greenup}{\textcolor{green}{$\uparrow$}}
            
\begin{document}

\title{FAST: Federated Active Learning with Foundation Models for Communication-efficient Sampling and Training}

\author{Haoyuan Li, Mathias Funk, Jindong Wang,~\IEEEmembership{Member,~IEEE}, Aaqib Saeed
\thanks{Haoyuan Li, Mathias Funk, and Aaqib Saeed are with the Department of Industrial Design, Eindhoven University of Technology, 5612 AZ Eindhoven, Netherlands}
\thanks{Jindong Wang is with the Department of Arts \& Sciences, College of William \& Mary, Williamsburg, Virginia 23185, USA.}
\thanks{Copyright (c) 20xx IEEE. Personal use of this material is permitted. However, permission to use this material for any other purposes must be obtained from the IEEE by sending a request to pubs-permissions@ieee.org.}
}
\markboth{Journal of \LaTeX\ Class Files,~Vol.~14, No.~8, August~2025}%
{Shell \MakeLowercase{\textit{et al.}}: A Sample Article Using IEEEtran.cls for IEEE Journals}

\IEEEpubid{0000--0000/00\$00.00~\copyright~2025 IEEE}

\maketitle

\begin{abstract}
Federated Active Learning (FAL) has emerged as a promising framework to leverage large quantities of unlabeled data across distributed clients while preserving data privacy. However, real-world deployments remain limited by high annotation costs and communication-intensive sampling processes, particularly in a cross-silo setting, when clients possess substantial local datasets. This paper addresses the crucial question: \emph{What is the best practice to reduce communication costs in human-in-the-loop learning with minimal annotator effort?} Existing FAL methods typically rely on iterative annotation processes that separate active sampling from federated updates, leading to multiple rounds of expensive communication and annotation. In response, we introduce \textbf{FAST}, a two-pass FAL framework that harnesses foundation models for weak labeling in a preliminary pass, followed by a refinement pass focused exclusively on the most uncertain samples. By leveraging representation knowledge from foundation models and integrating refinement steps into a streamlined workflow, \textbf{FAST} substantially reduces the overhead incurred by iterative active sampling.  Extensive experiments on diverse medical and natural image benchmarks demonstrate that \textbf{FAST} outperforms existing FAL methods by an average of 4.36\% while reducing communication rounds eightfold under a limited 5\% labeling budget.
\end{abstract}

\begin{IEEEkeywords}
federated learning, active learning, foundation model, communication efficiency.
\end{IEEEkeywords}

\section{Introduction}
\label{Introduction}
\IEEEPARstart{F}{ederated} Learning (FL) emerges as a key decentralized paradigm that enables edge clients (e.g., institutions or devices) to collaboratively train the unified model through global aggregation without compromising local data privacy \cite{mcmahan2017communication, li2020federated, kairouz2021advances}. In recent research, many FL approaches have been developed under the supervised learning setting, assuming that all training data on clients are fully annotated. However, in realistic scenarios, data are typically unlabeled, with only a very limited number of annotated instances. For instance, in the cross-silo scenario, a few organizations possess substantial datasets but face constraints in large-scale data annotation due to limited budgets, expertise, or time \cite{kairouz2021advances, liu2022privacy}. 

To tackle this challenge, recent studies \cite{deng2022fedal, cao2023knowledge, kim2023re, ahn2024federated, chen2024think} delve into the concept of federated active learning (FAL) which incorporate the active learning (AL) into the context of FL. AL aims to maximize model performance in situations with scarce labeled data and limited annotation budgets. It achieves this by iteratively selecting the most informative data instances for labeling by an oracle (i.e., a human annotator) based on specific query strategies. FAL bridges these two fields by incorporating active sampling steps during federated training rounds. Specifically, each client independently conducts active sampling on its local data, utilizing either the local model or the aggregated global model as a query selector to identify informative instances prior to local updates \cite{chen2024think, kim2023re, ahn2024federated}. After each AL iteration, local models are aggregated on the server to form a global model that can guide subsequent query selections. 

\IEEEpubidadjcol 
Recent advances in FAL have demonstrated significant benefits of AL in harnessing unlabeled data within the FL systems. While numerous studies have been proposed to address challenges posed by data heterogeneity in federated settings \cite{cao2023knowledge, kim2023re}, prior research has paid little attention to the additional communication costs incurred during federated active sampling. One major concern arises from the communication overhead caused by iterative local training on the updated labeled dataset during active sampling. This concern is particularly acute in cross-silo scenarios \cite{kairouz2021advances}, where each edge device (e.g., institution) holds a significant amount of data and requires extra communication support to achieve subpar global performance. Moreover, annotation costs in FL are inherently more expensive than in centralized AL due to the distributed and fragmented nature of the data across multiple clients, which requires further coordination and resource allocation. 

In our work, we aim to reduce the communication overhead during the FAL process with a limited annotation budget while achieving superior overall prediction performance of the global model. A critical challenge in FAL is the selection of a query selector for active sampling. \cite{ahn2024federated} investigate the discrepancy of utilizing a global or local-only model for active sampling and achieve robust performance by solely applying sampling strategies with the global model on IID data distribution. Nevertheless, \cite{chen2024think, cao2023knowledge} prove that the superiority of the query model depends on the heterogeneity of data distribution on the clients. Despite the advancement in exploring the utilization of query models, these methods all require sufficient active training rounds to iteratively improve the generalizability of client models as the feature extractor for selecting informative unlabeled samples. Instead of training the query model from scratch with the initial data pool from random sampling, we seek the applicability of foundation models in enhancing active sampling throughout the federated training process. Notably, previous research \cite{radford2021learning, zhai2023sigmoid, sun2023eva, oquab2023dinov2} on foundation models show that features learned from the foundation models are semantically organized in the representation space, providing robust and informative embeddings for downstream tasks. 
\IEEEpubidadjcol

Motivated by this, we introduce a two-pass \textbf{F}ederated \textbf{A}ctive learning framework with foundation models for communication-efficient \textbf{S}ampling and \textbf{T}raining, named \textbf{FAST}. In the initial pass, we leverage a frozen image encoder from a Vision-Language foundation model (e.g., SigLIP \cite{zhai2023sigmoid}) to perform weak labeling by selecting and prioritizing informative samples based on uncertainty estimates. This preliminary phase utilizes the semantic richness of foundation models to efficiently identify candidate data points for annotation. In the second pass, human oracles refine these weak labels to ensure labeling quality while operating under a limited labeling budget, thereby reducing communication overhead and minimizing the required human effort in the active sampling process. Our contributions are summarized as follows:
\begin{itemize}
    \item We investigate a challenging FAL scenario in which human annotation is costly and communication support is constrained, necessitating efficient strategies for both labeling and training.
    \item We propose a two-pass FAL framework to effectively utilize unlabeled data with minimal human intervention, achieving strong performance in a resource-efficient manner.
    \item We conduct extensive experiments on diverse benchmark datasets, covering both medical and natural images. Our results demonstrate that the proposed method outperforms existing approaches across various data distributions while reducing the required communication rounds by eightfold (8x) under only a 5\% labeling budget. 
\end{itemize}

\section{Related Work}
\subsection{Weakly Supervised Learning}
Weakly supervised learning (WSL) addresses scenarios where large portions of ground-truth labels are unavailable or limited. Based on the confidence of label availability, WSL is commonly divided into three paradigms: incomplete supervision, inexact supervision, and inaccurate supervision \cite{zhou2018brief, ren2023weakly}. Incomplete supervision involves abundant unlabeled instances and only a small subset of labeled data. This setting is often tackled either through active sampling (i.e., human intervention) or by exploiting semi-supervised learning with clustering or manifold assumptions \cite{dempster1977maximum, li2013convex, li2014towards}. Inexact supervision arises when only coarse-grained labels are provided, necessitating fine-grained instance-level identification via multi-instance learning algorithms \cite{settles2007multiple, wei2016empirical, wei2016scalable}. Lastly, inaccurate supervision denotes the presence of label noise \cite{frenay2013classification}, which is typically mitigated through label correction \cite{yi2019probabilistic, zheng2021meta, wu2021learning} or regularization-based robust training \cite{patrini2017making, hendrycks2018using, wang2019symmetric, lukasik2020does}. In this work, we focus on the incomplete supervision paradigm in the FL setting, where local datasets are largely unlabeled and distributed across multiple clients with minimal human intervention.

\subsection{Active learning}
Existing research in AL generally focuses on querying oracles to label the most informative data points, thereby minimizing labeling effort while maximizing model performance. The AL methods are typically divided into uncertainty-based, representativeness-based, and hybrid strategies. Uncertainty-based methods focus on samples with high aleatoric or epistemic uncertainty \cite{zhan2022comparative}, using metrics such as entropy, margin, or least confidence \cite{shannon1948mathematical, wang2014new, nguyen2019epistemic}. For example, BALD \cite{houlsby2011bayesian, gal2017deep, kirsch2019batchbald} seeks points maximizing mutual information between predictions and model parameters, while \cite{yoo2019learning} prioritizes samples expected to produce large errors.  Similarly, \cite{huang2021semi} employs Temporal Output Discrepancy to estimate uncertainty by measuring output discrepancies at different optimization steps. 

Representativeness-based methods aim to cover diverse regions of the input space to ensure broad decision boundaries. CoreSet \cite{sener2017active, geifman2017deep, caramalau2021sequential} addresses this by solving a k-center problem to create a representative core set. Additionally, clustering-based approaches, such as hierarchical clustering or self-organizing maps \cite{kutsuna2012active, citovsky2021batch}, and set coverage optimization \cite{urner2013plal, yang2017suggestive}, enhance representativeness and reduce redundancy in labeled data. In FL, clients engage in joint training of a global model while independently learning local models that can serve as query selectors. A naive way to adopt classical AL in FL is to apply local query sampling on individual clients. However, this approach faces significant challenges due to heterogeneous data distributions. In particular, local query selectors cannot fully leverage global knowledge, especially under non-IID conditions.

\subsection{Federated Active Learning}
Recent research has begun to investigate the applicability of AL within FL environments, where the scarcity of labeled client data constitutes a significant bottleneck for FL processes. Preliminary studies have focused on integrating AL into federated training by directly applying existing AL strategies to perform data annotation on client devices \cite{aussel2020combining, wu2022federated, alfalqi2023emergency, kong2023fedawr}. Nonetheless, conventional AL approaches are not specifically designed for decentralized data annotation, and numerous challenges remain unresolved. 

Unlike centralized AL, where the model independently selects samples for querying, FL enables clients to train the model collaboratively. In this context, \cite{deng2022fedal} explores the efficacy of global (F-AL) and local-only (S-AL) query selection in FL, revealing that F-AL effectively leverages inter-client collaboration to outperform S-AL. Further research on F-AL has sought to address the heterogeneity inherent in FL. \cite{cao2023knowledge} introduces a knowledge-aware method (KAFAL) to address the mismatch in sampling goals between local clients and the global model in non-IID federated settings. Similarly, \cite{kim2023re} proposes an innovative FAL sampling method (LoGo) that combines global and local model benefits to enhance inter-class diversity handling. \cite{chen2024think} integrates evidential learning with a Dirichlet-based model to handle uncertainty and improve data diversity, providing a robust solution for FAL in medical domains with domain shifts. 

Despite these advancements, communication overhead remains a core bottleneck for FAL. Each active sampling round typically involves additional local training and global aggregation steps, leading to high communication costs and substantial annotation efforts—particularly under cross-device FL with potentially millions of clients \cite{kairouz2021advances}. By contrast, our method focuses on the annotation process at the initial training stage, requiring only a limited labeling budget. We thus propose a communication-efficient FAL framework, \textbf{FAST}, that addresses both uncertainty and diversity in active sampling with minimal human effort.
\begin{figure*}
\centering\includegraphics[width=0.88\textwidth]{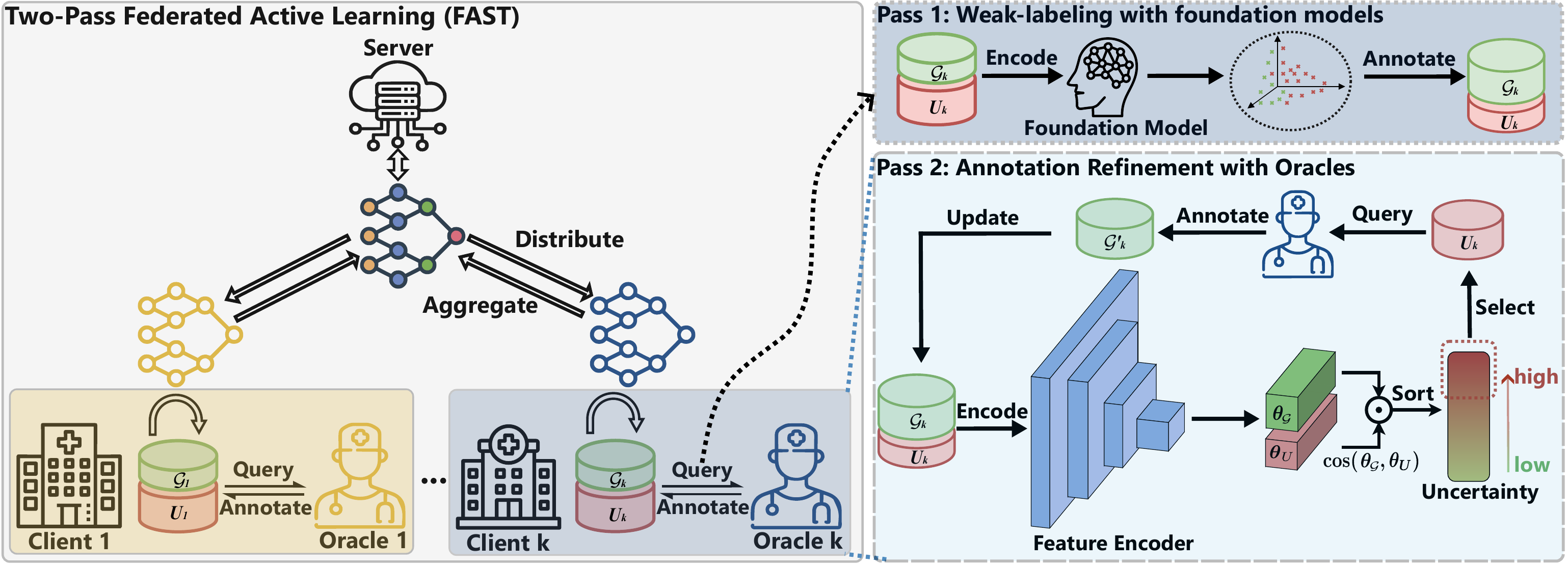}
\caption{Overview of \textbf{FAST}. \textbf{FAST} is a communication-efficient FAL framework that employs a two-pass labeling strategy. In the first pass, foundation models perform weak sampling to identify informative data points with minimal communication overhead. In the subsequent pass, human annotators refine the labeled ground truth dataset by validating and correcting the sampled labels, ensuring high-quality annotations.}
\label{fig:overview}
\vspace{-5mm}
\end{figure*}

\begin{algorithm}
\caption{\textbf{FAST}: Two-Pass Federated Active Learning}
\label{alg:fast}
\textbf{Data:} Local datasets $ D_k = \{ \mathcal{U}_k$, $\mathcal{G}_k ^{(0)} \}$.\\
\textbf{Input:} $K$ clients; $T$ federated rounds with $\tau$ local steps; Feature encoder $f(\cdot)$; Budget $B= \sum_{r=1}^{R}b$.\\
\textbf{Output:}  The target global model $w$.

\begin{algorithmic}[1]
\STATE \textbf{Initialize:} Server initializes global model with $w^{(0)}$.
\\
\texttt{======== Preliminary Pass ========}
\FOR{ client $k = 1, \dots, K$ (in parallel)}
    \STATE Feature encoding $ \mathbf{Z}_k \leftarrow f(\mathcal{U}_k) \cup f(\mathcal{G}_k^{(0)})$
    \STATE Perform label propagation on $\mathbf{Z}_{\mathcal{U}_k}$ to assign \emph{weak} 
    \\ \emph{labels} based on $\mathcal{G}_k^{(0)}$
    \FOR{class $c \in C$}
        \STATE Compute class similarity $\mathbf{s_{i,c}}$ for each weakly
        \\ labeled sample $x_i \in \mathcal{U}_k$ using Eq. (\ref{eq:class_sim})
    \ENDFOR
\ENDFOR

\texttt{========= Refinement Pass =========}
    \FOR{client $k = 1, \dots, K$ (in parallel)}
        \STATE Compute uncertainty score $u_i$ using Eq. (\ref{eq:comp_uncertain})
        \STATE Select top-$b$ samples with highest $u_i$ for each sample
        \\ $x_i \in \mathcal{U}_k$ for oracle annotation
        \STATE Update labeled set $\mathcal{G}_k \leftarrow \mathcal{G}_k \cup \mathcal{U}_k^{(b)}$
        \STATE Merge refined annotated data with labeled data $\mathcal{G}_k^{(0)}$
    \ENDFOR 

\texttt{======== Federated Training ========}
    \FOR{communication round $t =1, \dots, T$}
    \STATE \textbf{Client Update:} Distribute $w^{(t)}$ to clients in $K$.
    \FOR{ client $k = 1, \dots, K$ (in parallel)}
        \STATE Initialize local model $w_k^{(t)} \leftarrow w^{(t)}$.
        \FOR{$i = 0, \dots, \tau-1$}
            \STATE Perform local SGD updates on client $k$
            \\ $w_k^{(t+1)} \leftarrow w_k^{(t)} - \eta \nabla \mathcal{F}_k(w_k^{(t)})$.
        \ENDFOR
        \STATE Send updated $w_k^{(t+1)}$ back to server.
    \ENDFOR
    \STATE \textbf{Server Update:} Aggregate local models.
    \STATE Update global model $w^{(t+1)} \gets \frac{1}{|K|} \sum_{k \in K} w_k^{(t)}$
\ENDFOR
\STATE \textbf{Return} Target global model $w$.
\end{algorithmic}
\end{algorithm}

\section{Methodology}

\subsection{Problem Formulation}
Given a federated learning (FL) task involving $K$ clients, where each client $k$ possesses a local dataset $D_k$ stored on its device. The global dataset is the union of all local datasets, denoted as $D = \bigcup_{k=1}^{K} D_k$. The objective of FL is to collaboratively learn a global model by solving the following optimization problem in a distributed manner:
\begin{align}
\label{eq: FL-obj}
\min_{w} F(w) &\triangleq \frac{1}{K} \sum_{k=1}^{K} \mathcal{F}_k(w_k) \nonumber\\
&= \frac{1}{K} \sum_{k=1}^{K} \mathbb{E}_{(x, y) \sim \mathcal{D}_k}\bigl[\mathcal{F}_k(w_k; x_k, y_k)\bigr].
\vspace{-5mm}
\end{align}
where $w \in \mathbb{R}^d$ represents the global model parameters to be optimized. $F(w)$ is the global loss function aggregating the local losses from all clients, and $\mathcal{F}_k(w_k) = \mathbb{E}_{(x_k, y_k) \sim \mathcal{D}_k} \left[ \mathcal{F}_k(w_k; x_k, y_k) \right]$ is the expected risk over data distribution $\mathcal{D}_k$ at client $k$ corresponding to parameter vector $w_k$. $\mathcal{F}_k(w_k; x_k, y_k)$ denotes the loss incurred by the local model $w_k$ on data sample $(x_k, y_k)$ generated from the local data distribution of client $k$. In heterogeneous FL, data is distributed across clients in a non-IID manner, i.e., data distribution on each local client is distinct, for clients data $\{D_k, D_j\} \in D$, $\mathcal{D}_k \neq \mathcal{D}_j$. 

Previous studies typically \cite{mcmahan2017communication} solve Eq.\ref{eq: FL-obj} by iteratively updating the global model through local computations on each client and averaging client updates at the server. At communication round $t$, the server sends the current global model parameters $w^{(t)}$ to a selected subset of clients $\mathcal{K}_t \subseteq {1, 2, \dots, K}$. Each client $k \in \mathcal{K}_t$ initializes its local model with the received parameters, $w_k^{(t)} = w^{(t)}$, and performs $\tau$ steps of local stochastic gradient descent (SGD) on its local dataset $D_k$:
\begin{equation}
\label{eq: aggregate}
    w_k^{(t, i+1)} = w_k^{(t, i)} - \eta \nabla \mathcal{F}_k(w_k^{(t, i)}; \xi_k^{(t, i)}),
\end{equation}
where $\eta$ is the learning rate, $i = 0, 1, \dots, \tau - 1$, and $\xi_k^{(t, i)}$ denotes a mini-batch sampled from $D_k$. After local updates, clients send their updated local models $w_k^{(t)}$ back to the server. The server aggregates these models by computing an average to update the global model, $ w^{(t+1)} = \frac{1}{|\mathcal{K}_t|} \sum_{k \in \mathcal{K}_t} w_k^{(t)}$.

\subsection{Federated Active Sampling}  
AL aims to enhance model performance by iteratively querying and labeling the most informative and representative samples from an unlabeled dataset, under a limited annotation budget. In FAL, this process is adapted to the decentralized setting by executing local active sampling and federated training at each $r$ AL round, $r \subseteq {1, 2, \dots, R}$. 

We consider a standard FAL case, where, clients utilize the global model $w^{(t)}$ as the query selector for client-level sampling. During the active sampling phase, each client $k$ selects $b$ unlabeled samples from its local unlabeled dataset $\mathcal{U}_k$ using a predefined query strategy $\mathcal{A}(\cdot)$. At the first AL round, client $k$ randomly selects a small set of $b$ samples for annotation to form the initial labeled ground truth data $\mathcal{G}_k^0$:
\begin{equation} 
  \mathcal{G}_k^{(0)} = \mathcal{A}(\mathcal{U}_k, b) = \texttt{Random}(\mathcal{U}_k, b), \text{where}\ \mathcal{G}_k^{(0)} \in \mathcal{U}_k .
\end{equation}

In subsequent $R-1$ AL rounds, the query strategy $\mathcal{A}(\cdot)$ utilizes the aggregated global model $w^{(r)}$ from the previous round as the query selector to identify informative samples. The selected samples are then labeled and added to the labeled local dataset $\mathcal{G}_k$, while being removed from the unlabeled dataset $\mathcal{U}_k$
\begin{equation} 
\label{eq: data_update}
\mathcal{G}_k^{(r)} \leftarrow \mathcal{G}_k^{(r-1)} \cup \mathcal{A}(w_k^{(r)};\mathcal{U}_k, b), \quad \mathcal{U}_k \leftarrow \mathcal{U}_k \setminus \mathcal{A}(\mathcal{U}_k, b). 
\end{equation}

The active sampling process continues until the global labeling budget of $B$ is exhausted, ensuring that the total number of labeled samples across all clients does not exceed $B$.
\begin{equation} 
\label{eq: budget}
\sum_{k=1}^{K} |\mathcal{G}_k^{(r)}| \leq B, \quad \forall r. 
\end{equation}
After each active sampling step at round $r$, federated training is performed. Each client $k$ updates its local model $w_k^{(r)}$ by training on the updated labeled dataset $\mathcal{G}_k^{(r)}$, and sends their updated models to the server, which aggregates them to form the new global model $w^{t}$ as discussed in Eq. \ref{eq: aggregate}. Given $T$ federated training rounds, the overall federated rounds across $K$ clients is $R \times T \times K$.

\subsection{Two-Pass Federated Labeling}
We introduce \textbf{FAST}, a communication-efficient federated active learning framework grounded in a two-pass labeling strategy. In the preliminary pass, foundation models (e.g., vision or vision-language) generate preliminary labels based on their representation-based knowledge. This is followed by a refinement pass, where human annotators provide additional annotations to enhance label accuracy and reliability. Unlike previous FAL methods—which rely on iterative cycles of active sampling and federated training and thus incur significant communication overhead—FAST mitigates frequent client-server exchanges, substantially reducing overall communication costs.

In \textbf{FAST}, each client $k$ utilizes the frozen encoder from a pre-trained foundation model as a feature extractor $f(\cdot)$ to encode its local dataset $D_k$ into high-dimensional representations: $\mathbf{Z}_k = f(D_k), \ \mathbf{Z}_k \in \mathbb{R}^{d}$. Specifically, the unlabeled dataset $\mathcal{U}_k$ and the initial labeled dataset $\mathcal{G}_k^{(0)}$ are encoded:
\begin{equation} 
\mathbf{Z}_{\mathcal{U}_k} = f(\mathcal{U}_k), \ \mathbf{Z}_{\mathcal{G}_k} = f(\mathcal{G}_k^{(0)}). 
\end{equation}
To augment the labeled dataset with weak labels for the samples in $\mathcal{U}_k$, we perform label propagation on extracted representation $\mathbf{Z}_{\mathcal{U}_k}$ based on $k$-nearest neighbors in the embedding space. For each sample $x_i$ in unlabeled dataset $\mathcal{U}_k$, we assign the weak labels based on the majority vote of these neighbors with respect to $L_2$ distance in the initial labeled dataset $\mathcal{G}_k^{(0)}$. Next, we compute the cosine similarity between the embedding of each weakly labeled sample $x_i$ and the embeddings of all labeled samples in $\mathcal{G}_k^{(0)}$. For each class $c \in C$, we calculate the average cosine similarity $s_{i,c}$ between the embedding $\mathbf{z}_i$ of sample $x_i \in \mathcal{U}_k$ and the embeddings $\mathbf{z}_j$ of all labeled samples $x_j \in \mathcal{G}_{k,c}^{(0)}$:
\begin{equation}
\label{eq:class_sim}
s_{i,c} = \frac{1}{\left|\mathcal{G}_{k,c}^{(0)}\right|} \sum_{x_j \in \mathcal{G}_{k,c}^{(0)}} \frac{\mathbf{z}_i \cdot \mathbf{z}_j}{\|\mathbf{z}_i\| \|\mathbf{z}_j\|}, \quad \forall c \in C
\end{equation}
where $\mathcal{G}_{k,c}^{(0)}$ denotes the set of initial labeled samples of class $c$ at client $k$, and and $C$ represents the set of all classes. This process yields a prototype vector $\mathbf{s_i} = [s_{i,1}, s_{i,2},\ ..., s_{i,C}]$ for each weakly labeled sample $x_i$. The logits vector represents the average similarity of the sample to each class prototype in the labeled dataset, thereby capturing more nuanced relationships between the weakly labeled samples and the labeled data. 
We then utilize an uncertainty-based query function $\mathcal{A}(\cdot)$, such as entropy \cite{wang2014new}, on the softmax-normalized logits vector $\mathbf{s_i}$ to compute the uncertainty of each weakly labeled sample:
\begin{equation}
\label{eq:comp_uncertain}
\resizebox{\columnwidth}{!}{$
u_i = \mathcal{A}(\mathbf{s_i}) = -\sum_{c=1}^{C} \left( \frac{\exp(s_{i,c})}{\sum_{c'=1}^{C} \exp(s_{i,c'})} \right) \log \left( \frac{\exp(s_{i,c})}{\sum_{c''=1}^{C} \exp(s_{i,c''})} \right).
$}
\end{equation}
Samples with higher uncertainty scores $u_i$ are considered more informative. We rank the samples in $\mathcal{U}_k$ based on their uncertainty scores and select the top $b$ samples for annotation with the given labeling budget in Eq.\ref{eq: budget}. The newly annotated samples are added to the labeled dataset $\mathcal{G}_k$ and removed from the unlabeled dataset $\mathcal{U}_k$, as shown in Eq.\ref{eq: data_update}. Subsequently, these human-labeled samples are combined with the weakly labeled samples to form the final labeled dataset for the federated training process, eliminating the need for additional active sampling steps. We summarize the whole procedure of \textbf{FAST} approach in Algorithm \ref{alg:fast}.

\begin{figure*}[t]
 \centering
 \begin{subfigure}[t]{0.49\textwidth}
     \centering
     \includegraphics[width=\textwidth]{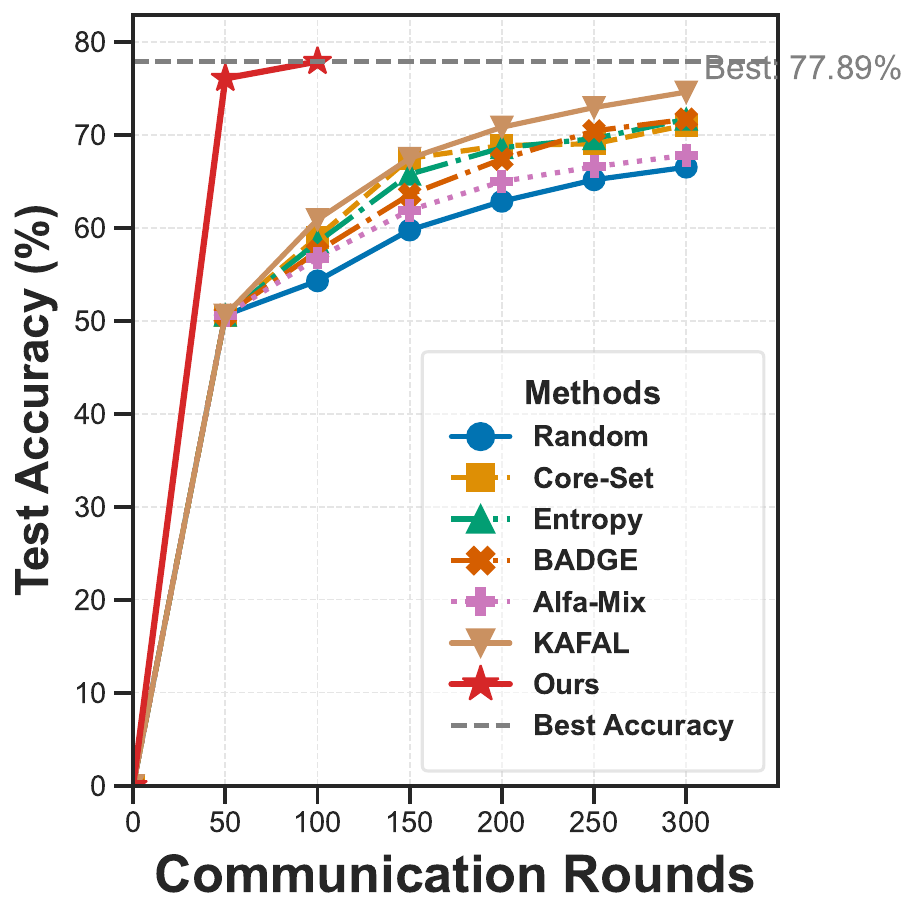}
     \caption{CIFAR-10}
     \label{com-cifar10}
 \end{subfigure}
 \hfill
 \begin{subfigure}[t]{0.49\textwidth}
     \centering
     \includegraphics[width=\textwidth]{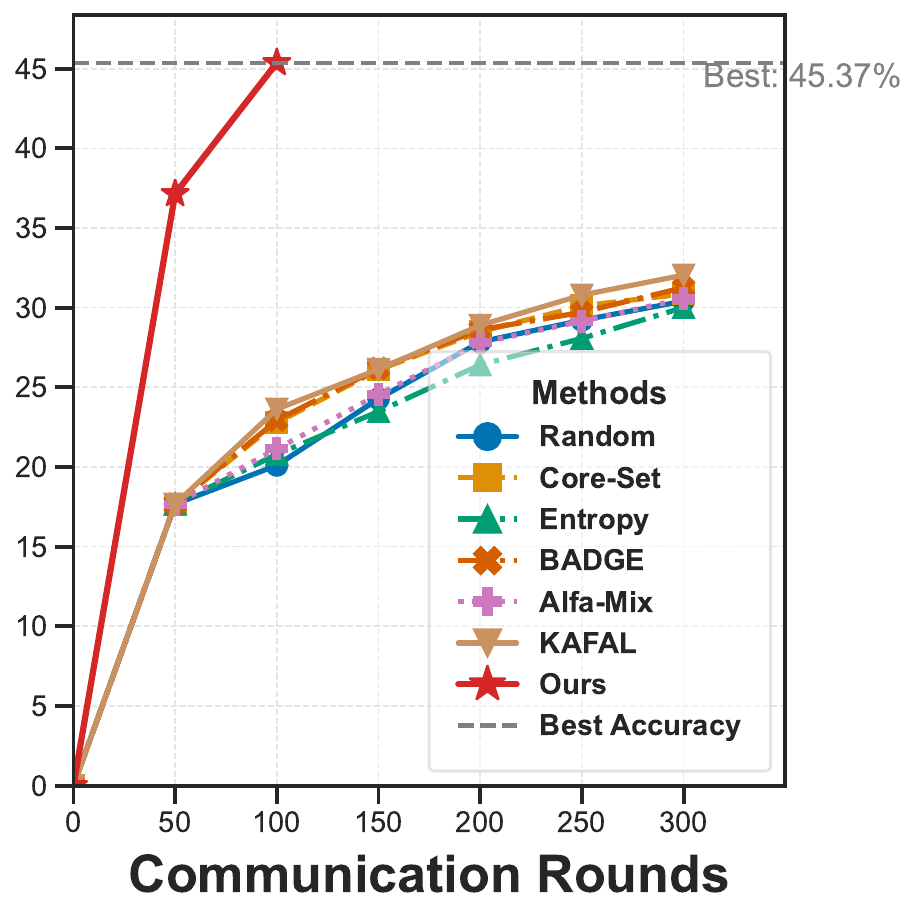}
     \caption{CIFAR-100}
     \label{com-cifar100}
 \end{subfigure}
 \caption{Experimental comparison of our method with existing approaches on CIFAR-10 and CIFAR-100 under a Non-IID data distribution. Other AL methods begin by randomly selecting 10\% of the initial data, followed by 50 communication rounds of training after each AL sampling step until reaching a 35\% labeling budget. In contrast, our method completes training at 100 rounds and achieves its highest performance (indicated by the grey line). Refer to Appendix Figure~\ref{exp:acc_vs_com_full} for results over the entire 300-round training process.}
 \label{exp:acc_vs_com}
 \vspace{-5mm}
\end{figure*}

\section{Experiments}

\subsection{Experimental Configuration}
\textbf{Datasets.} We evaluate our method primarily on image classification tasks spanning both natural and medical benchmark datasets. Specifically, we use four natural image datasets—CIFAR10/100 \cite{krizhevsky2009learning}, Tiny-ImageNet \cite{le2015tiny}, and SVHN \cite{netzer2011reading}—as well as two medical image datasets—PathMNIST and DermaMNIST \cite{medmnistv2}. To account for the inherent heterogeneity among clients, we consider three data distribution settings: IID, Non-IID, and heterogeneous inter-class diversity (i.e., variations in local class distributions) \cite{kim2023re}. As shown in Table \ref{exp:main_table}, we report the total labeling budget and the corresponding number of training rounds. Following existing FAL approaches, each client trains its local model from scratch and iteratively selects 5\% of the total dataset for annotation in each AL round, until reaching a predefined global labeling budget. We use a 20\% labeling budget for most of our experiments and ablation studies. To ensure fairness in labeling costs, we assume this global labeling budget is evenly shared among clients, such that each client queries the same number of samples per AL round. 
 
\textbf{Baselines.} We compare \textbf{FAST} with nine standard active learning (AL) strategies: Random, Entropy \cite{wang2014new}, Coreset \cite{sener2017active}, BADGE \cite{ash2019deep}, LL4AL \cite{yoo2019learning}, GCNAL \cite{caramalau2021sequential}, and ALFA-Mix \cite{parvaneh2022active}. Although originally designed for centralized AL, these strategies can be independently applied on either a global or a local model within a federated environment. In our experiments, we employ the global model as the query selector for active sampling. We further include two federated AL (FAL) strategies, KAFAL \cite{cao2023knowledge} and LoGo \cite{kim2023re}. For the Non-IID experiment in Figure \ref{exp:acc_vs_com}, we select KAFAL as it is specifically tailored for global heterogeneity problems. Regarding the experiment on IID datasets in Table \ref{exp:main_table}, we add LoGo as the baseline considering its focus on solving heterogeneous data from the client level.

\textbf{Implementation Settings.}
We implement our proposed \textbf{FAST} method in PyTorch using the Flower FL framework \cite{beutel2020flower}. As our primary federated learning (FL) strategy, we adopt FedAvg \cite{mcmahan2017communication}, and additionally evaluate on FedProx \cite{li2020federated} and FedNova \cite{wang2020tackling} to examine the robustness of \textbf{FAST} across different FL paradigms (see Table~\ref{tab:exp_fl_methods}). Our experiments primarily target cross-silo settings with full client participation, involving a total of $10$ clients. Each AL round spans $T=100$ federated communication rounds, and each client executes $\tau=5$ local stochastic gradient descent (SGD) steps per round. In alignment with prior work \cite{kim2023re, cao2023knowledge}, we employ a four-layer \emph{CNN} as our main model architecture and employ a \emph{ResNet-8} network for ablation studies on communication efficiency. We simulate the Non-IID data partitions by sampling from a Dirichlet distribution with a concentration parameter of $\alpha=0.1$, where smaller values of $\alpha$ indicate greater data heterogeneity across clients \cite{hsu2019measuring}. For the implementation of \textbf{FAST}, we initialize with 1\% of labeled data and employ a frozen SigLIP \cite{zhai2023sigmoid} as the foundation model for feature extraction and weak labeling in the two-pass process. We conducted all experiments on 2 NVIDIA A10 GPUs with Intel Xeon Gold 6342 CPUs (2.80GHz).

\subsection{Results}

\textbf{Comparison on heterogeneous inter-class diversity data.}  We first evaluate the performance of FAST in comparison with other baseline methods on datasets characterized by high levels of local heterogeneity. In this context, each client shares the same pool of classes but exhibits varying inter-class distributions. We deliberately fix FAST’s labeling budget at 5\% to underscore its cost-effectiveness (Table \ref{exp:main_table}). With just one AL round and 5\% labeled data, FAST matches or exceeds the accuracy of baselines that consume up to 40\% budget over multiple AL rounds. In contrast, conventional AL methods sample 5\% per round iteratively until they exhaust the same budget, incurring additional communication and computation. FAST achieves competitive performance through a single, one-shot active labeling round. This design choice highlights FAST’s significant efficiency advantages in reducing communication overhead.

Conventional AL methods aim to minimize labeling efforts by selecting a small subset of instances based on their informativeness across the entire dataset. However, in a decentralized setting where each local dataset maintains distinct class distributions, such imbalanced data partitions often lead to inconsistent knowledge sharing. Consequently, the selected samples may not be representative or sufficiently informative for all clients, thereby hindering the overall learning performance. As presented in Table \ref{exp:main_table}, we compare the performance of \textbf{FAST} with other existing AL strategies under a one-shot setting, wherein only a single active sampling round is conducted. We observe that \textbf{FAST} outperforms all baseline methods even within the constraints of this one-shot scenario. Notably, in this experiment, the server exhausts 5\% of the labeling budget per round until reaching the total budget limit. 

\begin{table*}[t]
    \centering
    \caption{Test accuracy comparison of various Active Learning (AL) strategies across multiple datasets. We evaluate FAST in a one-shot (i.e., a single AL round) setting, where each AL round is followed by 100 Federated Learning (FL) rounds, resulting in a total of $R_{\text{FL}} = R_{\text{AL}} \times 100$. The labeling budget denotes the percentage of data allocated for labeling, with each AL round querying 5\% of the unlabeled samples for annotation.}
    \label{exp:main_table}
    \resizebox{0.9\textwidth}{!}{%
        \sisetup{
            table-number-alignment = center,
            detect-mode,
            detect-family,
            detect-inline-family = math,
            round-mode = places,
            round-precision = 2
        }
        \begin{tabular}{
            lcS[table-format=2.2]S[table-format=2.2]S[table-format=2.2]S[table-format=2.2]cc
        }
            \toprule
            \multicolumn{1}{c}{\textbf{Method}} & 
            \multicolumn{1}{c}{\textbf{R\(_{AL}\)}} & 
            \multicolumn{1}{c}{\textbf{CIFAR-10}} & 
            \multicolumn{1}{c}{\textbf{SVHN}} & 
            \multicolumn{1}{c}{\textbf{PathMNIST}} & 
            \multicolumn{1}{c}{\textbf{DermaMNIST}} & 
            \multicolumn{1}{c}{\textbf{R\(_{FL}\)}} & 
            \multicolumn{1}{c}{\textbf{Budget}} \\
            \midrule
            \multirow{2}{*}{Random} 
                & 4 & 64.19 & 80.90 & 68.41 & 71.70 & 400 & 20\% \\
                & 8 & 69.07 & 84.22 & 73.76 & 72.66 & 800 & 40\% \\ 
            \midrule
            \multirow{2}{*}{Entropy \cite{wang2014new}}
                & 4 & 64.02 & 82.08 & 71.54 & 72.49 & 400 & 20\% \\
                & 8 & 69.12 & 85.88 & 75.91 & 73.02 & 800 & 40\% \\ 
            \midrule
            \multirow{2}{*}{Coreset \cite{sener2017active}}
                & 4 & 64.66 & 80.94 & 74.84 & 72.02 & 400 & 20\% \\
                & 8 & 69.43 & 83.81 & 76.85 & 72.34 & 800 & 40\% \\ 
            \midrule
            \multirow{2}{*}{BADGE \cite{ash2019deep}}
                & 4 & 65.12 & 82.81 & 72.21 & 72.59 & 400 & 20\% \\
                & 8 & 69.57 & 85.89 & 75.53 & 73.23 & 800 & 40\% \\ 
            \midrule
            \multirow{2}{*}{GCNAL \cite{caramalau2021sequential}}
                & 4 & 65.40 & 82.05 & 75.51 & 72.01 & 400 & 20\% \\
                & 8 & 70.05 & 85.09 & 78.13 & 73.07 & 800 & 40\% \\ 
            \midrule
            \multirow{2}{*}{ALFA-Mix \cite{parvaneh2022active}}
                & 4 & 65.45 & 83.02 & 73.34 & 72.39 & 400 & 20\% \\
                & 8 & 69.87 & 86.05 & 76.31 & 73.27 & 800 & 40\% \\ 
            \midrule
            \multirow{2}{*}{LoGo \cite{kim2023re}}
                & 4 & 66.50 & 83.46 & 76.32 & 72.61 & 400 & 20\% \\
                & 8 & 71.70 & 86.02 & 79.51 & 73.33 & 800 & 40\% \\
            \midrule
            \textbf{Ours} & \textbf{1} & \textbf{77.14} & \textbf{87.91} & \textbf{88.48} & \textbf{74.37} & \textbf{100} & \textbf{5\%} \\ 
            \bottomrule
        \end{tabular}
    }
\end{table*}

In \textbf{FAST}, each client shares their representation-based knowledge with other clients without revealing the raw local dataset, thereby enabling the server's query selector to address imbalanced class distributions from a global perspective. By fully exploiting the comprehensive information of the unlabeled dataset, \textbf{FAST} is able to achieve superior performance after the first AL round.

\textbf{FAST under a fixed communication budget with Non-IID data distribution.} We assess the effectiveness of FAST in a non-IID data setting by distributing client data according to a Dirichlet distribution with a concentration parameter of $\alpha=0.1$, thereby inducing high data heterogeneity across all clients. Figure \ref{exp:acc_vs_com} illustrates the comparative convergence rates of the global model on CIFAR-10 and CIFAR-100, where the grey line denotes the optimal performance achieved by all AL methods. To evaluate the communication efficiency of our approach, we conduct a total of $T=300$ FL communication rounds across all methods. For the baseline methods, we initialize the process with 10\% of labeled data at the beginning of the first 50 FL rounds. In the subsequent federated training phases, the server queries 5\% of unlabeled instances for human annotation every 50 rounds until the total communication budget is exhausted. 

In contrast, FAST employs a two-pass active sampling process at the onset of the AL phase to utilize the predefined global labeling budget without necessitating further oracle participation. As depicted in Figure \ref{exp:acc_vs_com}, FAST achieves superior global model performance by the $100_{th}$ FL round without depleting the allocated communication budget. These results demonstrate that our method enables the server to efficiently train a high-performing global model within limited communication resources in realistic scenarios.

\textbf{Impact of two-pass active sampling on Foundation Models with Linear Probing.} In this experiment, we evaluate the efficacy of a two-pass sampling strategy within \textbf{FAST} by integrating a foundation model as the backbone during training. Instead of training client-specific models from scratch, we employ linear probing on the client side using only a limited labeled dataset. To systematically analyze the contribution of each component, we decompose the training process into four distinct elements: \circled{1} Linear Probing, \circled{2} Weak Labeling, \circled{3} Active Learning, and \circled{4} Random Sampling. 

\begin{table*}[!ht]
\centering
\resizebox{0.7\textwidth}{!}{%
    \renewcommand{\arraystretch}{1.2} 
    \begin{tabular}{cccc|cccc}
    \hline
    \textbf{\circled{1}} & \textbf{\circled{2}} & \textbf{\circled{3}} & \textbf{\circled{4}} & \textbf{CIFAR-10} & \textbf{CIFAR-100} & \textbf{Tiny-ImageNet} & \textbf{PathMNIST} \\
    \hline
    \ding{51} & \ding{51} & \ding{51} &  & \textbf{96.04} & \textbf{60.83} & \textbf{54.41} & \textbf{86.67} \\
    \hline
    \ding{51} & \ding{51} &  & \ding{51} & 95.31 & 58.94 & 52.95 & 82.33 \\
    \ding{51} & \ding{51} &  &           & 94.47 & 53.56 & 46.92 & 75.84 \\
    \ding{51} &  &  & \ding{51}          & 94.53 & 52.84 & 47.79 & 74.12 \\
    \ding{51} &  &  &                    & 80.43 & 5.61 & 1.60 & 49.89 \\
    \hline
    \end{tabular}
}
\caption{Effects of training components: \circlednumber{1}~Linear Probing, \circlednumber{2}~Preliminary Pass, \circlednumber{3}~Refinement Pass, \circlednumber{4}~Random. We train with a limited 1\% of initial labeled data across all 10 clients for 100 FL rounds. The labeling budget is 20\%.}
\label{exp:component_analysis}
\vspace{-5mm}
\end{table*}

Table \ref{exp:component_analysis} illustrates the performance outcomes of various component combinations across multiple datasets under a fixed labeling budget of 20\%, encompassing 100 FL rounds distributed among 10 clients, with an initial training dataset comprising 1\% of labeled data for all clients. Specifically, we consider five different scenarios to examine the efficacy of the two-pass mechanism in \textbf{FAST}, where the combination of the first three components (\circled{1}, \circled{2}, and \circled{3}) represents the integration of \textbf{FAST} into linear probing. In Table \ref{exp:component_analysis}, the configuration employing the two-pass sampling strategy (\circled{1}, \circled{2}, \circled{3}) achieves superior performance compared to the configurations that only implement preliminary labeling (\circled{1}, \circled{2}) and those that omit oracle refinement phase (\circled{1}, \circled{2}, \circled{4}). This demonstrates the critical role of human refinement during the FAL process in enhancing model performance. Notably, we observe significantly lower performance when directly applying linear probing with the foundation model on the initial labeled data without any further AL operations (\circled{1} only). These findings collectively highlight that the two-pass active sampling mechanism in \textbf{FAST} not only maximizes the utility of the limited labeling budget but also fosters effective knowledge sharing across heterogeneous clients, thereby achieving superior global model performance with constrained communication resources. 

Moreover, our analysis in Table \ref{exp:component_analysis} decomposes the FAST framework into distinct components for qualitative evaluation. Notably, we observe that applying uncertainty sampling (\circled{1}+\circled{2}+\circled{3}) consistently outperforms random sampling in the refinement pass (\circled{1}+\circled{2}+\circled{4}) across various datasets. These results empirically validate the effectiveness of uncertainty sampling within the second pass of our framework.

\subsection{Ablation Studies}

\textbf{Analysis on computational and communications overhead in FAST.} 
In FL, communication overhead is most commonly measured either by the number of rounds or by the total volume of parameter transmissions (i.e., model uploads and downloads) \cite{luping2019cmfl, al2021reducing, morell2022optimising}. As shown in Table I and Figure 2 of the main paper, FAST achieves higher accuracy within 100 FL rounds while reducing the number of communication rounds by a factor of eight. Table \ref{tab:comm-overhead} reports the aggregate parameter‐transmission cost of each method. In FAST’s preliminary stage, a small set of labeled embeddings is uploaded to and redistributed from the server for weak labeling, incurring an initial communication cost. However, by leveraging representation‐based labeling and minimal human annotation, FAST greatly reduces the required number of FL rounds and thus the overall communication volume. In contrast, conventional FAL methods incur identical model‐aggregation and distribution costs in every round, and their extensive active‐learning cycles result in comparable or higher total communication overhead.

Furthermore, we compare computational costs in terms of wall‐clock time (Table \ref{tab:comm-overhead}). The only additional expense in FAST’s preliminary stage is encoding local data via the foundation model’s encoder for uncertainty estimation. By contrast, LoGo \cite{kim2023re} performs both macro‐ and micro‐level informativeness evaluations using local and global models in each active‐learning round, imposing a substantially higher runtime burden. As the table shows, on CIFAR-10 LoGo incurs approximately 2310 s more wall‐clock time than random sampling over the same number of rounds.\\

\begin{table*}[ht]
  \centering
  \caption{Comparison of classification accuracy (\%), communication cost (MB), and wall-clock time (s) for FAST, LoGo, and random sampling on CIFAR-10, SVHN, and PathMNIST under the IID setting. We apply FedAvg on a four-layer CNN classifier (with a size of 0.45 MB on the CIFAR-10 dataset). FAST is evaluated with one AL round (100 FL rounds), whereas Random and LoGo use 8 AL rounds (i.e., 800 FL rounds).}
  \label{tab:comm-overhead}
  \resizebox{\textwidth}{!}{%
    \begin{tabular}{l *{9}{c}}
      \toprule
      Method 
        & \multicolumn{3}{c}{\textbf{CIFAR-10} }
        & \multicolumn{3}{c}{\textbf{CIFAR-100}} 
        & \multicolumn{3}{c}{\textbf{SVHN}} \\
      \cmidrule(lr){2-4} \cmidrule(lr){5-7} \cmidrule(lr){8-10}
        & Acc.\,(\%) & Comm. Cost (MB) & Walltime (s)
        & Acc.\,(\%) & Comm. Cost (MB) & Walltime (s)
        & Acc.\,(\%) & Comm. Cost (MB) & Walltime (s) \\
      \midrule
      Random     & 69.14 & 7090.94 & 30398.95 & 32.67 & 8502.69 & 37258.75 & 85.47 & 6969.79 & 58064.42 \\
      LoGo     & 71.92 & 7090.94 & 32709.14 & 34.27 & 8502.69 & 39268.24 & 87.08 & 6969.79 & 61638.97 \\
      Ours   & ~~~\textbf{77.16}~\greenup & ~~~~~~~\textbf{902.54}~\textcolor{green}{(87.27\%)}~\greendown & ~~~~~\textbf{7342.52}~\textcolor{green}{(76.73\%)}~\greendown & ~~~\textbf{41.94}~\greenup & ~~~\textbf{1079.56}~\textcolor{green}{(87.30\%)}~\greendown & ~~~~\textbf{15104.38}~\textcolor{green}{(60.53\%)}~\greendown & ~~~\textbf{88.79}~\greenup & ~~~~\textbf{896.72}~\textcolor{green}{(87.13\%)}~\greendown & ~~~\textbf{19178.83}~\textcolor{green}{(67.96\%)}~\greendown \\
      \bottomrule
    \end{tabular}%
  }
\end{table*}

\textbf{Ablation on different federated learning strategies.} We investigate the impact of various FL strategies on the performance of \textbf{FAST} under a fixed labeling budget of 20\%. Table~\ref{tab:exp_fl_methods} reports the accuracy across five benchmark datasets. Notably, FedNova offers marginal yet consistent improvements over FedAvg and FedProx on most datasets, indicating that \textbf{FAST} is compatible with advanced FL aggregation strategies and can further support heterogeneous scenarios. These findings confirm the robustness of \textbf{FAST} under different federated aggregation schemes.
\begin{table}[ht!]
    \centering
    \caption{Performance of Our Method Across Different Federated Learning Strategies with 20\% Labeling Budget}
    \label{tab:exp_fl_methods}
    \resizebox{0.495\textwidth}{!}{%
        \begin{tabular}{l *{5}{S[table-format=2.2]}}
            \toprule
            \textbf{Strategy} & \textbf{CIFAR-10} & \textbf{CIFAR-100} & \textbf{SVHN} & \textbf{PathMNIST} & \textbf{Tiny-ImageNet} \\
            \midrule
            \textbf{FedAvg}   & 73.81 & 34.77 & 86.27 & 84.64 & 26.03 \\
            \textbf{FedProx}  & 73.63 & 32.84 & 83.19 & 85.36 & 25.90 \\
            \textbf{FedNova}  & 74.12 & 36.60 & 87.12 & 87.92 & 28.30 \\
            \bottomrule
        \end{tabular} 
    }

\end{table}

\textbf{Effect of Different Foundation Model Selections on FAST.} We next evaluate how the choice of foundation model for the preliminary pass in \textbf{FAST} influences its overall performance. Specifically, we compare three vision-language models—CLIP, EvaCLIP, and SigLIP—along with an image-specific model, DINOv2, using a pre-trained ResNet-50 as the baseline. As shown in \Cref{tab:federated_active_learning}, EvaCLIP consistently achieves the highest accuracy across all datasets, followed closely by SigLIP and DINOv2. This underscores the importance of rich representation knowledge for enhancing weak labeling quality in the preliminary pass. Furthermore, the results suggest that leveraging expressive embeddings can significantly improve active sampling outcomes, even under constrained annotation budgets.

\begin{table}[h]
    \centering
    \caption{Performance Comparison of Our Methods with Different Foundation Models}
    \label{tab:federated_active_learning}
    \resizebox{0.49\textwidth}{!}{%
        \begin{tabular}{l *{5}{S[table-format=2.2, table-space-text-post = {\,\%}]}}
            \toprule
            \textbf{Dataset} & \textbf{ResNet-50} & \textbf{CLIP} & \textbf{Eva-CLIP} & \textbf{SigLIP} & \textbf{DINOv2} \\
            \midrule
            CIFAR-10      & 77.86 & 83.81 & 85.98 & 84.87 & 85.34 \\
            CIFAR-100     & 28.86 & 38.32 & 53.27 & 50.41 & 50.38 \\
            PathMNIST     & 82.67 & 87.73 & 91.04 & 88.79 & 89.19 \\
            \bottomrule
        \end{tabular}
    }
\end{table}

\begin{table}[ht]
    \centering
    \caption{Performance of Our Method with Varying Labeling Budgets. Training with the FedAvg strategy using a CNN-4 model, 10 clients, 100 rounds.}
    \label{tab:performance_scaling}
    \resizebox{0.35\textwidth}{!}{%
        \sisetup{
            table-number-alignment = center,
            detect-mode,
            detect-family,
            detect-inline-family = math,
            round-mode = places,
            round-precision = 2,
            table-space-text-post = {\,\%} 
        }
        \setlength{\tabcolsep}{6pt} 
        \begin{tabular}{@{}l *{4}{S}@{}}
            \toprule
            \textbf{Dataset} & \multicolumn{4}{c}{\textbf{Labeling Budget}} \\
            \cmidrule(lr){2-5}
             & {0\%} & {5\%} & {40\%} & {80\%} \\
            \midrule
            CIFAR-10  & 75.92 & 76.73 & 77.24 & 77.48 \\
            CIFAR-100 & 31.33 & 33.34 & 39.65 & 44.27 \\
            PathMNIST & 73.16 & 75.89 & 82.28 & 85.46 \\
            \bottomrule
        \end{tabular}
    }
\end{table}
\textbf{Ablation on Labeling Budget.} To assess the scalability of FAST with respect to the labeling budget, we evaluate its performance under varying labeling budgets ranging from 0\% to 80\%. Table \ref{tab:performance_scaling} illustrates the accuracy of FAST across various datasets as the labeling budget increases. The results demonstrate a positive correlation between the labeling budget and model accuracy, with significant performance improvements observed as the budget increases. For instance, on CIFAR-10, accuracy improves from 75.92\% at 0\% budget to 77.48\% at 80\% budget. Similar trends are observed across CIFAR-100 and Path-MNIST, indicating the effectiveness of FAST in leveraging additional unlabeled data to enhance model performance under constrained labeling budgets.

\section{Conclusion}
In this paper, we introduced a two-pass FAL framework, \textbf{FAST}, designed to address the critical challenges of limited annotation budgets and communication-intensive sampling processes in FAL. Our approach leverages robust representation-based knowledge from foundation models to efficiently query informative unlabeled data for annotation, thereby minimizing human effort and communication overhead. Extensive experiments on diverse vision datasets demonstrate that FAST consistently outperforms existing FAL methods in terms of both predictive performance and communication cost. These findings underscore the potential of leveraging foundation models to enhance FAL under realistic resource constraints. Future directions include exploring more sophisticated query strategies within \textbf{FAST} and quantifying weak labeling quality. Developing an additional filtering mechanism after the two-pass labeling process to enable label correction prior to final human annotation, thereby further enhancing performance and communication efficiency.

\section*{Acknowledgments}
This work was partially supported by the NGF AiNed Fellowship Grant of A.S. We acknowledge the use of the Dutch SURF Research cloud to run the experiments presented in this paper.
\appendix
\section*{Additional Experiments}

\begin{figure*}[t]
    \centering
    \begin{subfigure}[t]{0.49\textwidth}
        \centering
        \includegraphics[width=\textwidth]{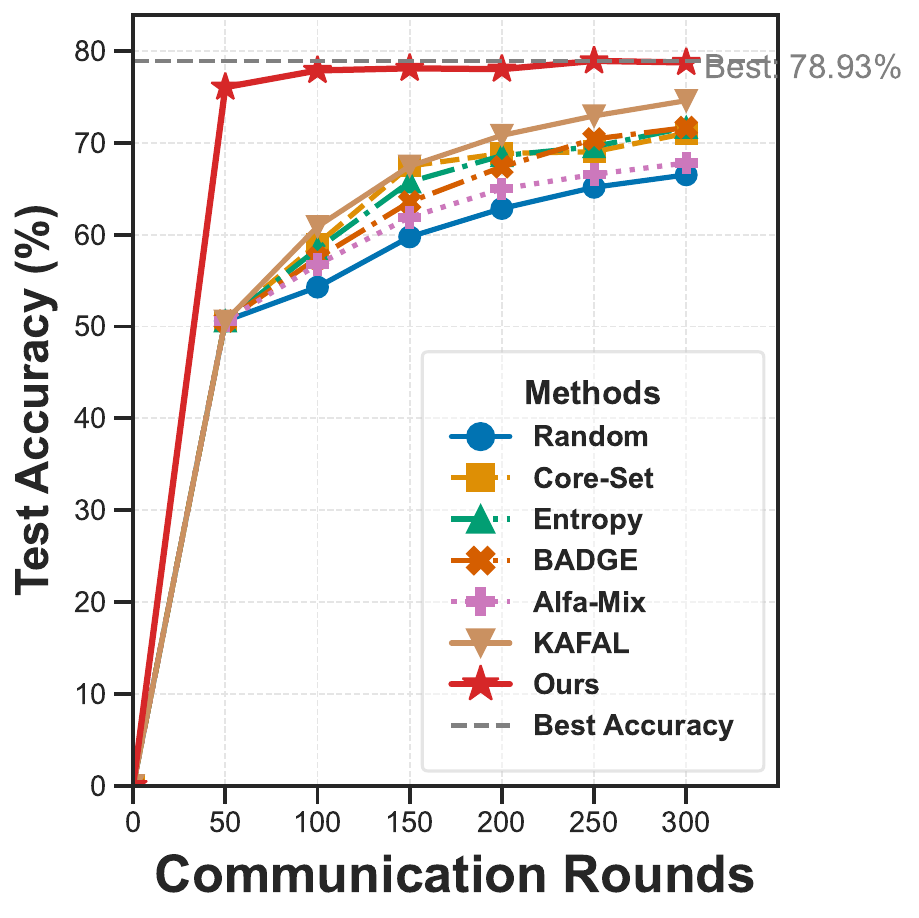}
        \caption{CIFAR-10}
        \label{com-cifar10_full}
    \end{subfigure}
    \hfill
    \begin{subfigure}[t]{0.49\textwidth}
        \centering
        \includegraphics[width=\textwidth]{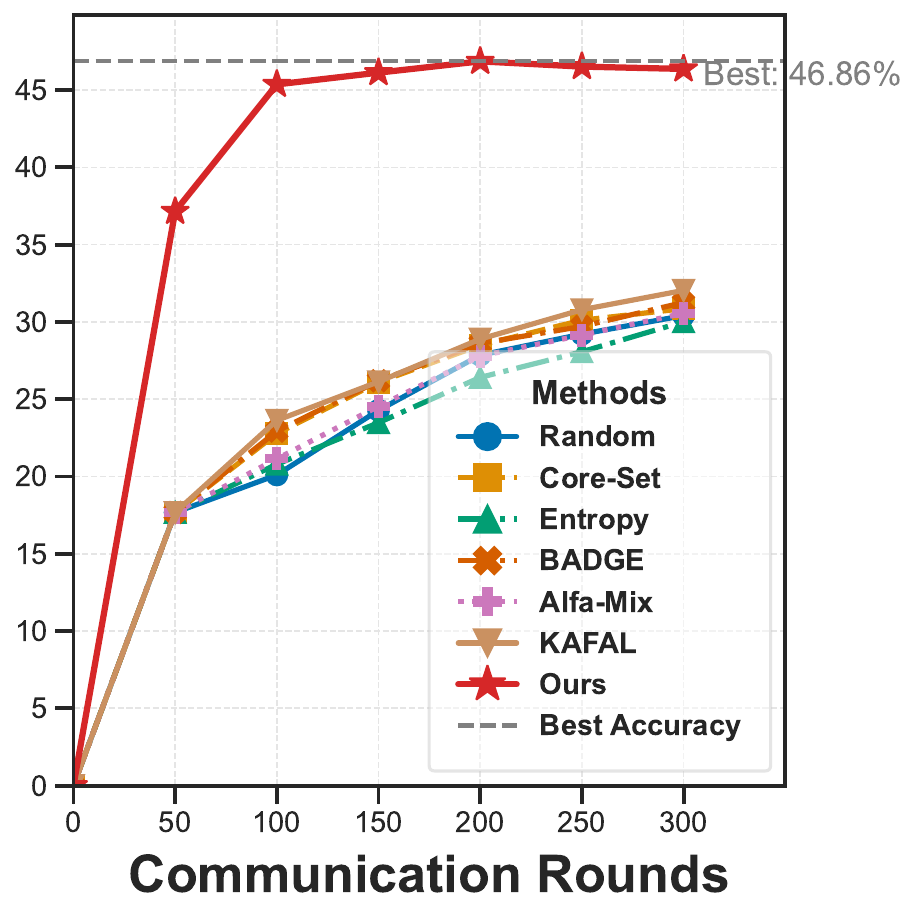}
        \caption{CIFAR-100}
        \label{com-cifar100_full}
    \end{subfigure}
    \caption{Experimental comparison of our method with existing approaches on the CIFAR-10 and CIFAR-100 datasets under a Non-IID data distribution. For other active learning (AL) methods, the process begins by randomly selecting 10\% of the initial data, followed by training with 50 communication rounds after each AL sampling step until a labeling budget of 35\% is reached.}
    \label{exp:acc_vs_com_full}
\end{figure*}

\begin{table*}[ht]
    \centering
    \caption{Performance of uncertainty sampling strategies on weak-labeled data across various datasets. Training with 10 clients for 100 rounds, utilizing a 4-layer CNN network. Evaluating with FedAvg.}
    \label{tab:uncertainty_sampling_performance}
    \small 
    \resizebox{0.9\textwidth}{!}{%
        \begin{tabular}{l *{5}{S[table-format=2.2, table-space-text-post = {\,\%}]}}
            \toprule
            \textbf{Dataset} & \textbf{Norm-Based} & \textbf{Entropy-Based} & \textbf{Least Confidence} & \textbf{Smallest Margin} & \textbf{Largest Margin} \\
            \midrule
            CIFAR-10      & 73.81 & 73.79 & 73.62 & 74.14 & 73.90 \\
            CIFAR-100     & 34.77 & 35.55 & 35.49 & 35.72 & 35.25 \\
            PathMNIST     & 84.64 & 85.43 & 85.29 & 84.85 & 85.70 \\
            Tiny-ImageNet & 28.37 & 29.18 & 28.89 & 28.72 & 28.91 \\
            \midrule
            \textbf{Average} & 55.40 & 55.74 & 55.82 & 55.86 & 55.94 \\
            \bottomrule
        \end{tabular}
    }
\end{table*}

\subsection*{Effect of Varying the Number of Clients on FAST.} 
\begin{figure}[h]
\centering\includegraphics[width=0.4\textwidth]{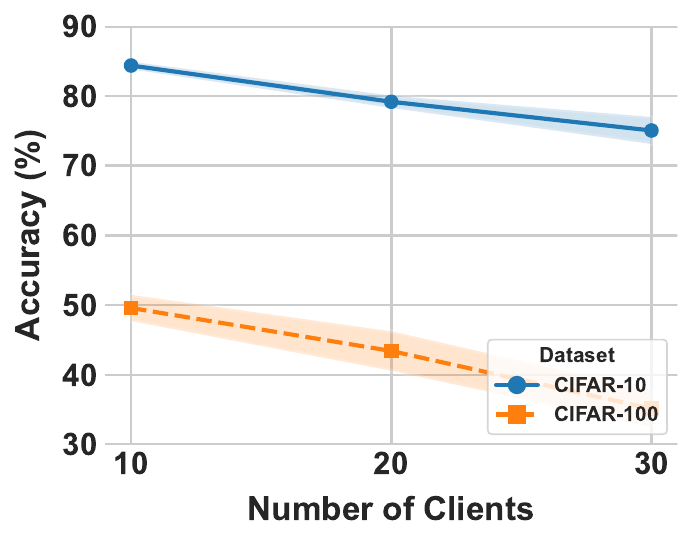}
\label{ablation:client_amount}
\caption{Performance of \textbf{FAST} across 10, 20, and 30 clients on CIFAR-10/100 under FedAvg with 150 FL rounds.}
\vspace{-5mm}
\end{figure}

We evaluate \textbf{FAST} with 10, 20, and 30 clients on CIFAR-10 and CIFAR-100 to assess its scalability and robustness. As shown in \Cref{ablation:client_amount}, the test accuracy decreases smoothly as the client count increases, indicating that more federated training rounds may be needed for convergence. Nonetheless, even under a limited annotation budget, \textbf{FAST} maintains strong performance without significant degradation, demonstrating its stability in larger federated learning clusters. 

\subsection*{FAST under a fixed communication budget with Non-IID data distribution.}
In Figure \ref{exp:acc_vs_com}, we showed FAST’s rapid convergence within the first 100 communication rounds. For completeness, Figure \ref{exp:acc_vs_com_full} presents extended results up to 300 rounds under the same Non-IID setting. As before, each method starts with a 10\% initial labeling and continues AL sampling until reaching 35\% of the labeling budget. 

\subsection*{Evaluating Uncertainty Strategies for Prototype-Based Weak Labeling.} 
We evaluate several uncertainty-based query strategies: norm-based, entropy-based, least confidence, smallest margin, and largest margin—applied to the prototype vectors computed for each weakly labeled sample. As summarized in \Cref{tab:uncertainty_sampling_performance}, the results are generally comparable across different datasets, suggesting that the prototype-based logits capture the key uncertainty information leveraged by a variety of query strategies. This underscores the effectiveness of the prototype representation in identifying highly uncertain samples for human refinement.


\newpage
\bibliography{refer}
\bibliographystyle{IEEEtran}


%


 




\vfill
\end{document}